\documentclass{article}
\usepackage{spconf,amsmath,graphicx,amssymb}
\usepackage{booktabs}
\usepackage{color}
\usepackage{hyperref}
\usepackage{overpic}
\usepackage{comment}
\usepackage{cite}

\newcommand{\R}{\mathbb R}
\newcommand{\ve}[1]{\mathbf#1}  
\newcommand{\m}[1]{\mathbf#1}  

\title{MLF-SC: Incorporating multi-layer features to sparse coding\\
for anomaly detection}
\name{Ryuji~Imamura \quad Kohei~Azuma \quad Atsushi~Hanamoto \quad Atsunori~Kanemura}
\address{LeapMind Inc., Tokyo, Japan\\
\{rimamura, azuma, hanamoto, atsu-kan\}@leapmind.io}
\begin{document}
\abovedisplayskip=2pt plus 1pt minus 1pt
\belowdisplayskip=2pt plus 1pt minus 1pt

%
\maketitle
\begin{abstract}
Anomalies in images occur in various scales from a small hole on a carpet to a large stain.  However, anomaly detection based on sparse coding, one of the widely used anomaly detection methods, has an issue in dealing with anomalies that are out of the patch size employed to sparsely represent images.
A large anomaly can be considered normal if seen in a small scale, but it is not easy to determine a single scale (patch size) that works well for all images.
Then, we propose to incorporate multi-scale features to sparse coding and improve the performance of anomaly detection. The proposed method, multi-layer feature sparse coding (MLF-SC), employs a neural network for feature extraction, and feature maps from intermediate layers of the network are given to sparse coding, whereas the standard sparse-coding-based anomaly detection method directly works on given images. We show that MLF-SC outperforms state-of-the-art anomaly detection methods including those employing deep learning. Our target data are the texture categories of the MVTec Anomaly Detection (MVTec AD) dataset, which is a modern benchmark dataset consisting of images from the real world.
Our idea can be a simple and practical option to deal with practical data.
\end{abstract}
\begin{keywords}
Anomaly detection, sparse coding, textures, VGG16
\end{keywords}
\section{Introduction}
\label{sec:intro}

Sparse coding for anomaly detection uses the reconstruction errors of local patches as the measure of anomalousness and has successfully been applied for discovering anomalies and novelties from practical data ~\cite{sparse_coding_work1, sparse_coding_work2, sparse_coding_for_texture,Cong:2011}. Sparse coding reconstructs all of the small patches of an image by combining a small number of bases in a given dictionary that is generated only from normal images. Since the dictionary contains only normal images, it is expected that the reconstruction errors for normal images are small whereas those for anomalies are relatively large. Therefore, we threshold the reconstruction error to detect anomalies. 

\begin{figure}[tbp]
  \vspace{-8mm}
  \centering
  \includegraphics[width=0.96\linewidth]{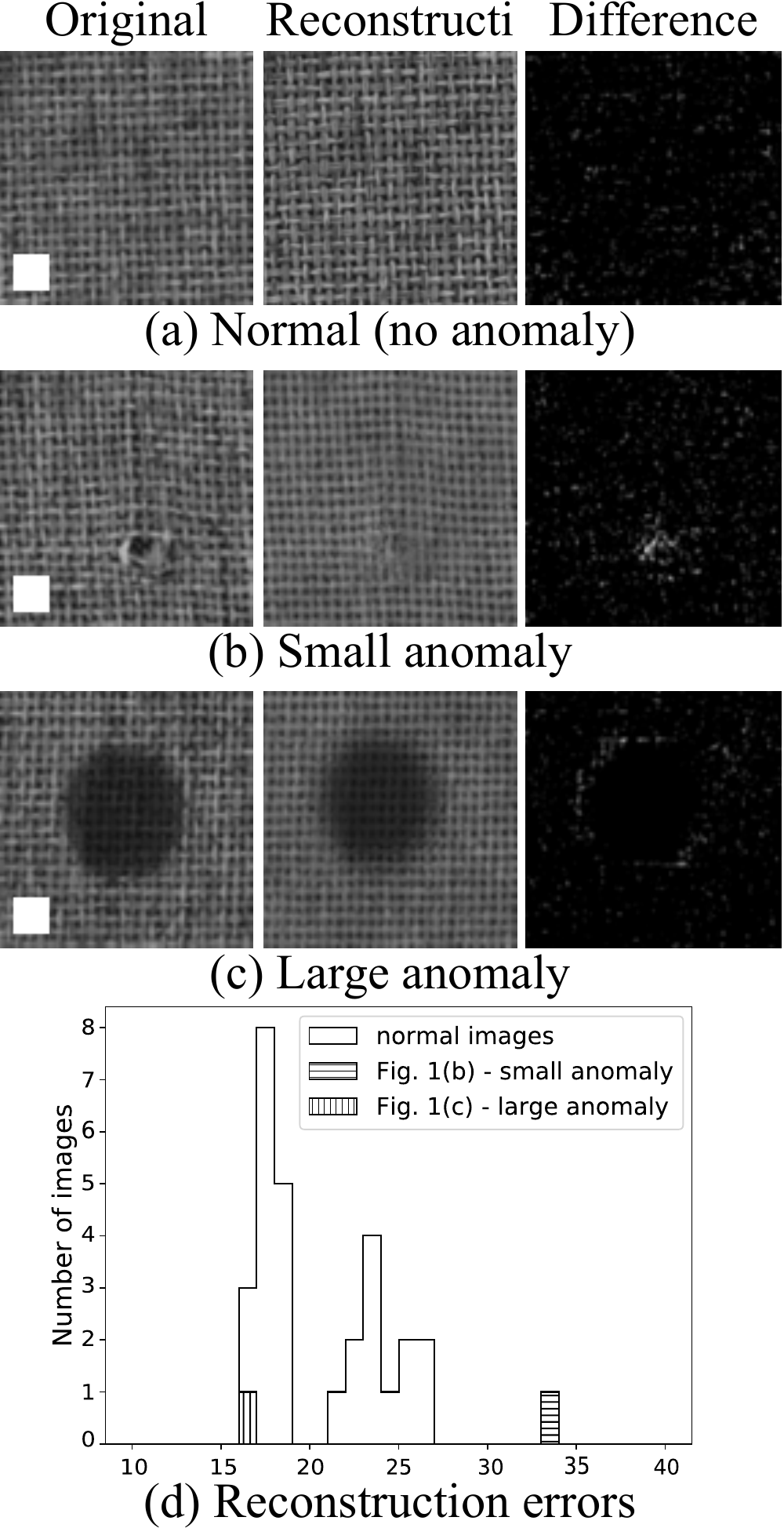}
  \vspace{-3mm}
  \caption{When sparse coding fails to detect anomalies. An original image, its reconstruction with sparse coding, and their difference are shown on one row for each of (a)~normal, (b)~small anomaly, and (c)~large anomaly images. The white box in the original images is the patch size for sparse coding. (d)~The histogram of the reconstruction errors of normal images and those for the two anomalous images. The original images are from the MVTec AD dataset~\cite{mvtec_add}.}
  \label{fig:when_sc_fails}
\end{figure}

A drawback of anomaly detection based on sparse coding is that its scale and ability to detect anomalies are limited to the patch size, and therefore large anomalies that span multiple patches can be difficult to detect.  This drawback can be understood through the illustrative examples in Fig.~\ref{fig:when_sc_fails}.  The normal image shown in Fig.~\ref{fig:when_sc_fails}(a) is reconstructed well and the reconstruction errors for normal images are distributed between 15 and 30 (Fig.~\ref{fig:when_sc_fails}(d)).  The small-anomaly image shown in Fig.~\ref{fig:when_sc_fails}(b) is not well represented by sparse coding and its reconstructed error is 34, which is out of the normal range. However, the large-anomaly image shown in Fig.~\ref{fig:when_sc_fails}(c) yields errors only at the edges of the anomaly and the resulting reconstruction error is as low as 16, which is not detected to be anomalous.
Therefore, the performance of anomaly detection with sparse coding depends on the patch size.

In this paper, we propose to use a multi-layer neural network as a feature extractor for sparse-coding-based anomaly detection and describe a novel anomaly detection method, which we term multi-layer feature sparse coding (MLF-SC).  Since the multi-layer neural network has multiple scales in its intermediate representations, MLF-SC can cope with the scale-related drawback of sparse coding. We use the VGG16 network~\cite{Simonyan2015vgg} for multi-scale feature extraction and achieve superior anomaly detection performance over the state-of-the-art and classical methods on the texture data from the MVTec AD dataset~\cite{mvtec_add}.

\section{Related Works}
\label{sec:related_works}

The current state-of-the-art methods for anomaly detection in texture images are autoencoder~\cite{ae_anomaly_detection}, AnoGAN~\cite{anogan}, and the CNN feature dictionary~\cite{cnn_feature_dictionaly}, all of which employ multi-layer architecture of deep neural networks.  Autoencoder is a bottleneck-architecture network, and trained to reconstruct images with only normal images. A trained autoencoder is not good at reconstructing images that have different characteristics from training data, then we can detect anomalies if the reconstruction errors are large; this is the one of the basic principles of anomaly detection commonly employed also by other methods.  AnoGAN employs a generative adversarial network (GAN)~\cite{gan} instead of autoencoder. The CNN feature dictionary method extracts features of an image by using a pretrained model, which are clustered by $K$-means.

The MVTec Anomaly Detection (MVTec AD) dataset~\cite{mvtec_add} is a recently created with the aim to be a standard dataset in the anomaly detection task, like MNIST~\cite{mnist} or ImageNet~\cite{imagenet} in the image classification task.
MVTec AD consists of five texture categories (including carpet and tile) and ten object categories (e.g.\ bottles and screws), where real images are collected.  Many anomaly detection studies have made an unnatural use of datasets like MNIST or CIFAR-10~\cite{cifar10} that have been created for classification purposes, by treating one class as anomaly.  However, MVTec AD contains real anomaly images and is more practical than previously employed datasets.

Anomaly detection methods based on sparse coding~\cite{sparse_coding_work1,sparse_coding_work2,sparse_coding_for_texture,Cong:2011} are not recognized as state-of-the-art, and not selected by Bergmann et al.~\cite{mvtec_add} for evaluating the MVTec AD dataset.

\begin{figure*}[tbp]
  \includegraphics[width=0.9\linewidth]{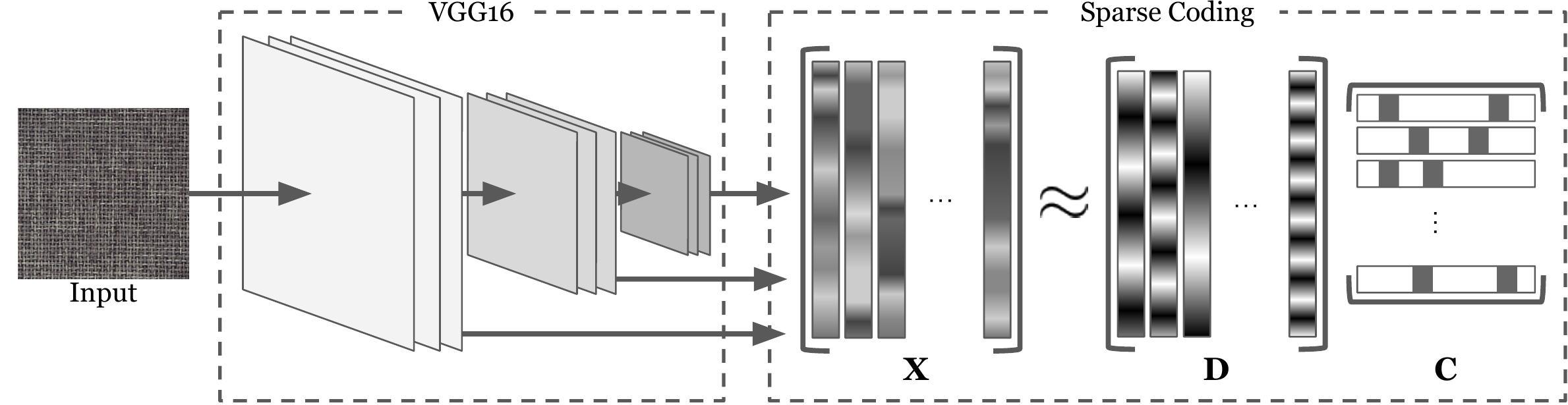}
  \vspace{-4mm}
  \caption{The data processing flow of MLF-SC. The input image is supplied to the VGG16 network, whose intermediate feature maps are given to sparse coding.}
  \label{fig:arch_of_proposed_method}
\end{figure*}

\section{Proposed Method---MLF-SC}
\label{sec:proposed_method}

This section describes the proposed method, MLF-SC, which detects anomalies by combining sparse coding and the features from intermediate layers of VGG16~\cite{Simonyan2015vgg}.  Fig.~\ref{fig:arch_of_proposed_method} shows the data flow in MLF-SC.

\subsection{Sparse Coding for Anomaly Detection}

The basic principle of anomaly detection with sparse coding is to reconstruct a given image using a dictionary learned only from normal images and threshold the reconstruction error~\cite{sparse_coding_work2, sparse_coding_for_texture,Cong:2011}. Since the dictionary is learned to represent only normal images, the reconstruction error is expected to be small if the given input image is normal, and the reconstruction error is expected to be large if the given image has anomalies that cannot be captured by the normal dictionary.

Sparse coding~\cite{img_denoising_by_sparse_coding,Elad:2010aa} expresses input signals, i.e.\ patches extracted from the input image, $\ve x_i \in \R^M$ ($i = 1, \dotsc, I$) by a linear combination of a few bases (the patch size is $\sqrt M \times \sqrt M$). We prepare $N\ (> M)$ basis vectors $\ve d_n$ ($n = 1, \dotsc, N$) and construct a dictionary matrix $\m D = [\ve d_1, \dotsc, \ve d_N] \in \R^{M \times N}$. Sparse coefficients $\ve c_i \in \R^N$ are found by solving the following minimization problem.
\begin{equation}
 \min_{(\ve c_i)_{i=1}^I} 
  \sum_{i=1}^I \biggl(\frac12\lVert\m x_i - \m D\ve c_i\rVert_2^2 + \alpha \lVert\ve c_i\rVert_1\biggr)
\label{eq:l1_sparse_coding}
\end{equation}
where the $\ell_1$ regularization makes $\ve c_i$ sparse and $\alpha$ is the regularization parameter.

To learn a dictionary from normal data, we extracted patches $\ve y_j \in \R^M$ ($j = 1, \dotsc, J$) from normal images and solve the following optimization problem~\cite{Elad:2010aa,Mairal:2010}.
\begin{equation}
\min_{\m D, (\ve c_j)_{j = 1}^J} \sum_{j=1}^J\biggl(\lVert \ve y_j - \m D\ve c_i \rVert_2^2 + \alpha \lVert\ve c_i\rVert_1\biggr)
\text{ s.t. } \forall n\lVert\ve d_n\rVert_2 = 1,
\label{eq:l1_dict_learn}
\end{equation}
where the unit-norm constraint on the basis vectors is to avoid the multiplication indeterminacy between $\ve d_n$ and $c_{in}$.

\subsection{MLF-SC}

MLF-SC follows the same principle of thresholding the reconstruction error for anomaly detection, but the difference from other sparse-coding-based methods is that MLF-SC uses the feature maps from intermediate layers of a deep neural network as input signals for sparse coding instead of images patches from the original, raw image. 

To obtain deep features, we employ VGG16~\cite{Simonyan2015vgg}, which is a deep neural network pretrained with the ImageNet dataset, winning the ILSVR 2014 Competition at the second place in the classification task. VGG16 is still widely used in feature extraction and transfer learning because the intermediate features from VGG16 contain rich information about an input image~\cite{style_transfer,imagenet_trained_cnn}.

\begin{figure}[tbp]
  \centering
  \includegraphics[width=0.6\columnwidth]{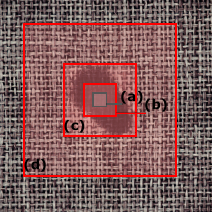}
  \vspace{-4mm}
  \caption{How MLF-SC incorporates multi-scale processing. The rectangles show the receptive fields when the patch size is 16${}\times{}$ 16. Region~(a) is the receptive field of conventional sparse coding, and regions~(b), (c), and (d) are the receptive fields at the 4th, 7th, and 10th convolutional layers of VGG16, respectively.}
  \label{fig:receptive_fields_comparison}
\end{figure}

MLF-SC uses the outputs from the 4th, 7th, and 10th convolution layers before the 2nd, 3rd, and 4th max pooling layer of VGG16 as input signals for sparse coding, so that MLF-SC can detect multi-scale anomalies. In this case, ``image'' patches should be called ``feature'' patches. As shown in Fig.~\ref{fig:receptive_fields_comparison}, if we fix the patch size $\sqrt M \times \sqrt M$ of the input signals, i.e.\ the feature patches, the size of the corresponding image patch (the receptive field) on the original image becomes larger for deeper layers because of a repetition of convolution operations; this is the source of the multi-scale processing of MLF-SC.

\begin{figure*}[tbp]
\begin{center}
\begin{minipage}{0.16\hsize}
\begin{center}
\includegraphics[width=\hsize]{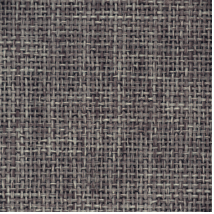}
non-anomaly image from {\it Carpet}
\end{center}
\end{minipage}
\begin{minipage}{0.16\hsize}
\begin{center}
\includegraphics[width=\hsize]{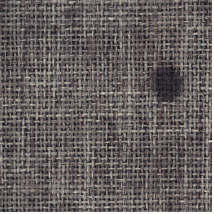}
(a) anomaly image (color)
\end{center}
\end{minipage}
\begin{minipage}{0.16\hsize}
\begin{center}
\includegraphics[width=\hsize]{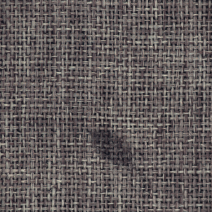}
(b) anomaly image (color)
\end{center}
\end{minipage}
\begin{minipage}{0.16\hsize}
\begin{center}
\includegraphics[width=\hsize]{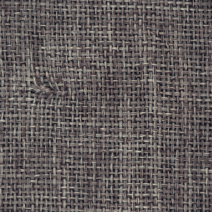}
(c) anomaly image (cut)
\end{center}
\end{minipage}
\begin{minipage}{0.16\hsize}
\begin{center}
\includegraphics[width=\hsize]{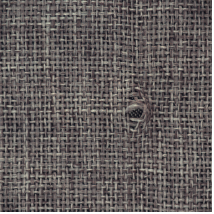}
(d) anomaly image (hole)
\end{center}
\end{minipage} \\

\begin{minipage}{0.3\hsize}
\begin{center}
\includegraphics[clip, trim=20 0 40 30, width=\hsize]{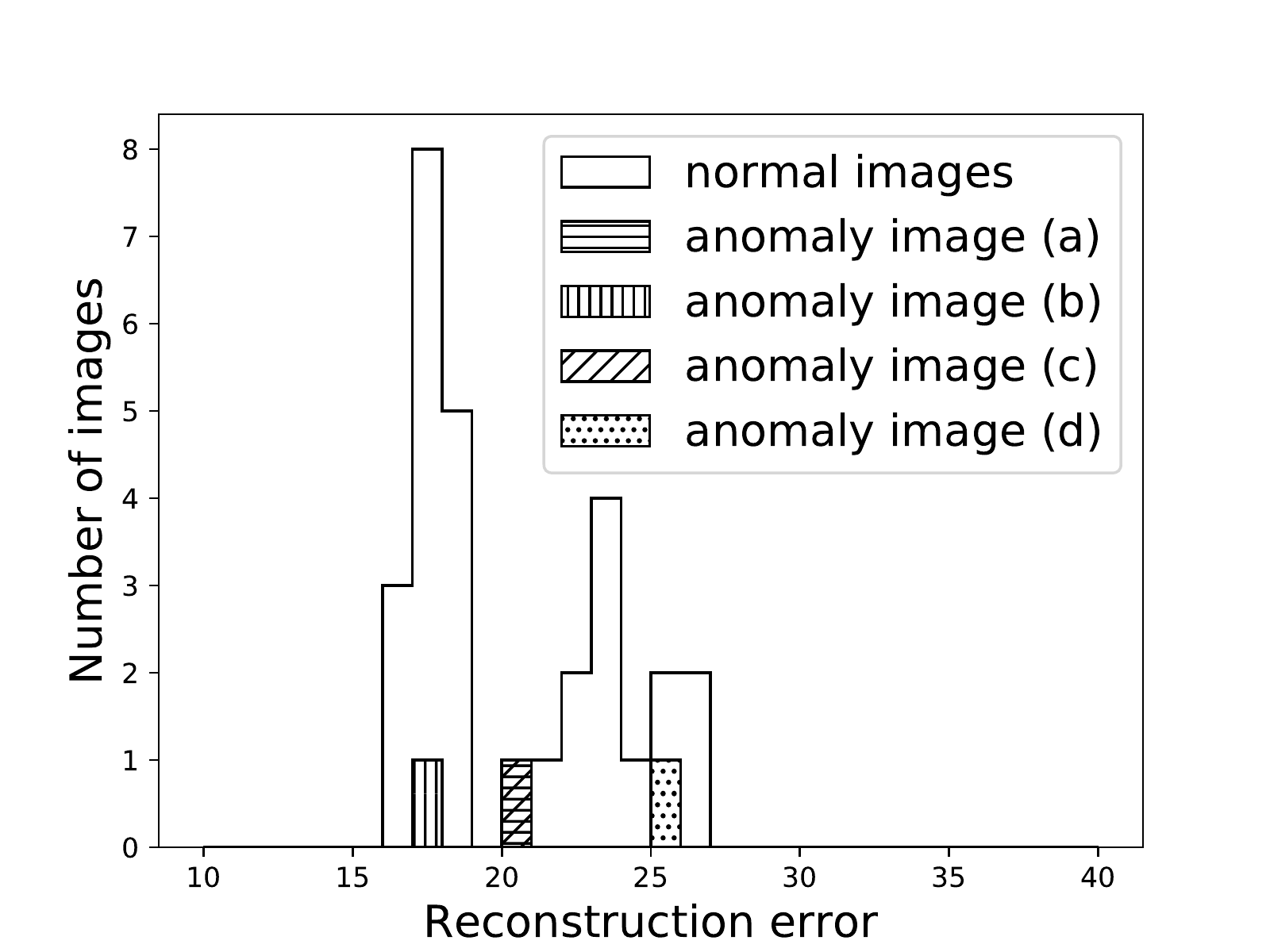}
Input Layer (Original Image)
\end{center}
\end{minipage}
\begin{minipage}{0.3\hsize}
\begin{center}
\includegraphics[clip, trim=20 0 40 30, width=\hsize]{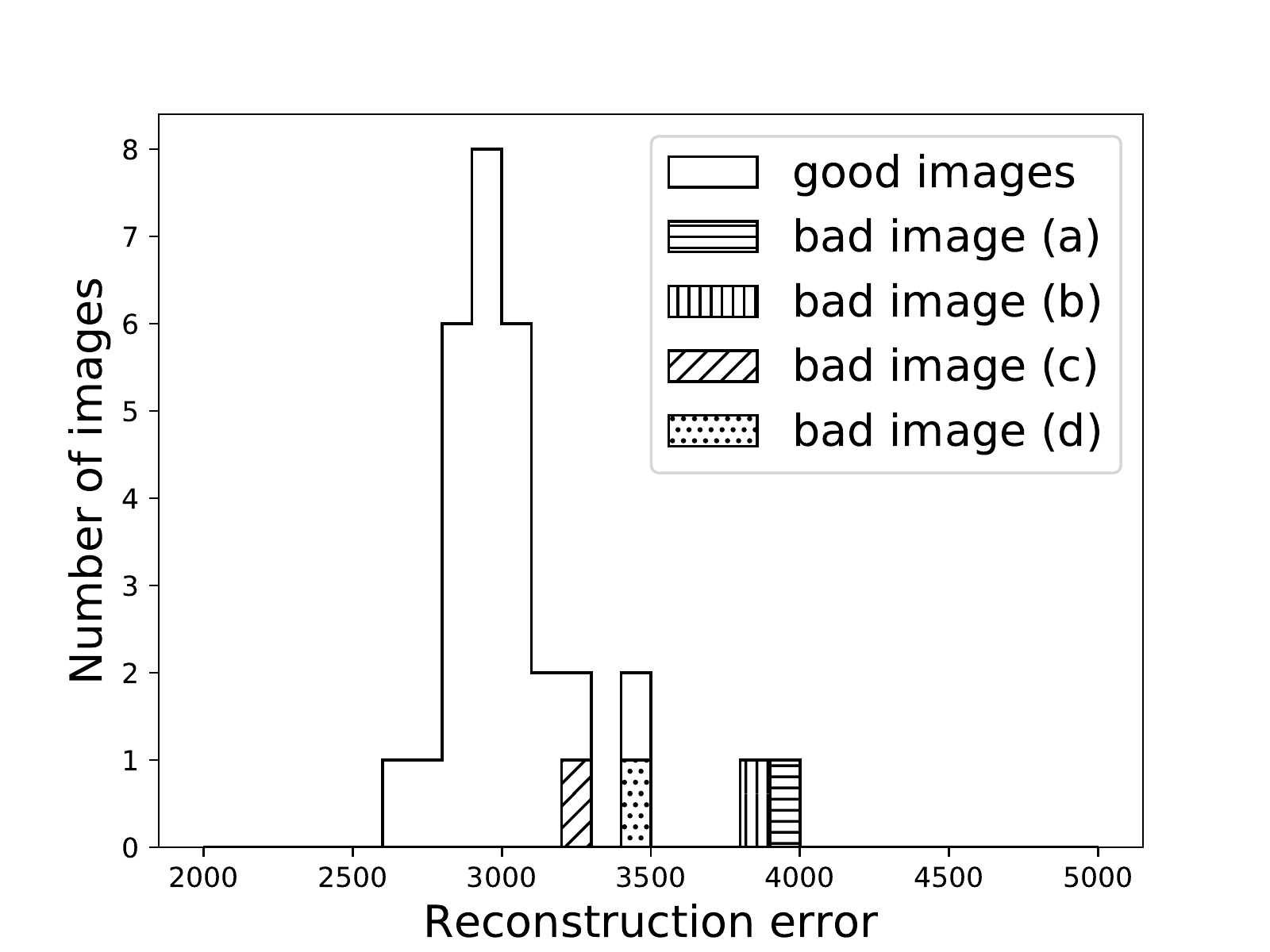}
Shallow Layer
\end{center}
\end{minipage}
\begin{minipage}{0.3\hsize}
\begin{center}
\includegraphics[clip, trim=20 0 40 30, width=\hsize]{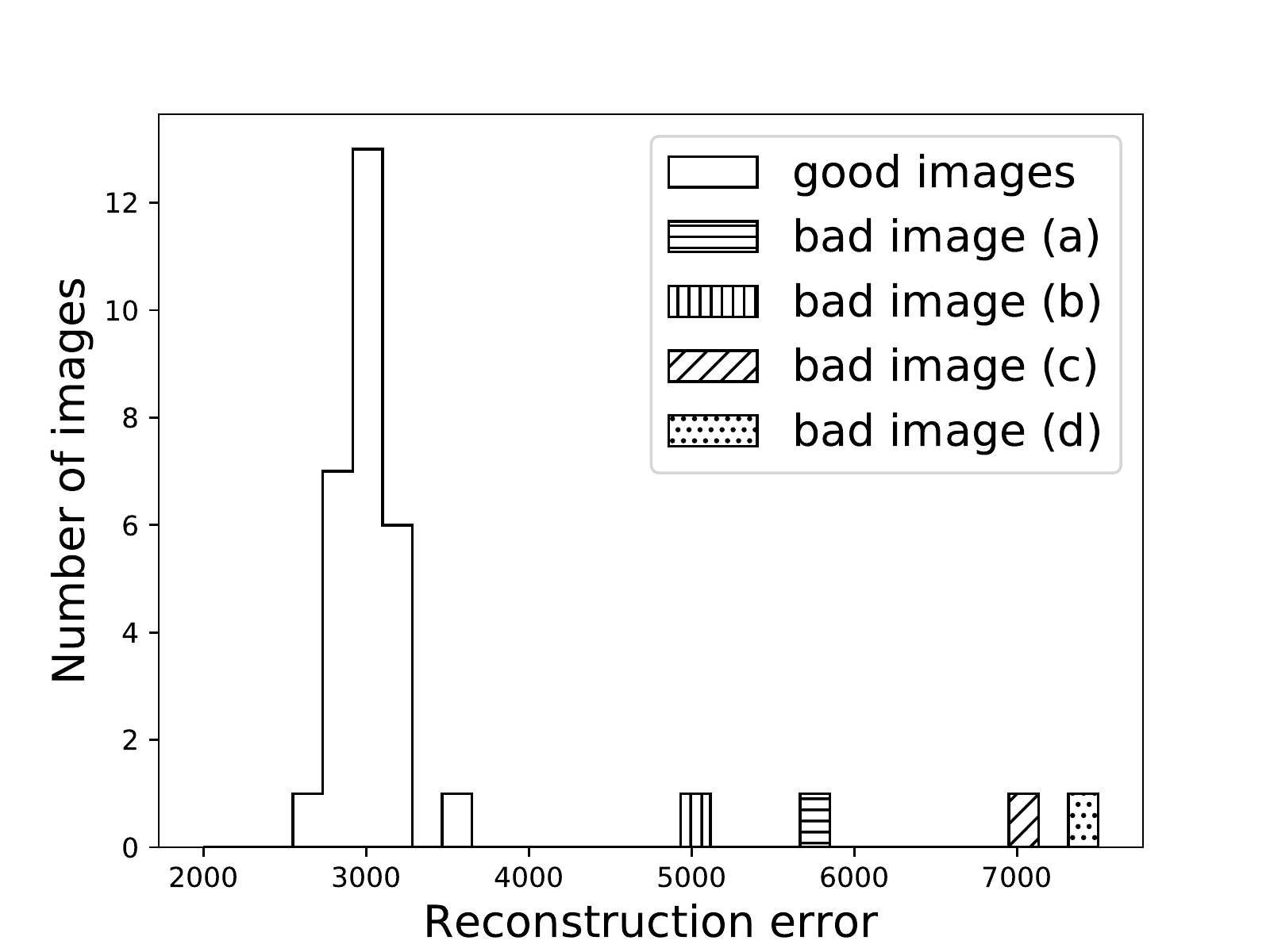}
Deep Layer
\end{center}
\end{minipage}
\end{center}
\vspace{-5mm}
\caption{Input images and histograms of reconstruction errors in each layer.}
\label{fig:example_images_and_histograms_for_discussion}
\end{figure*}

\begin{table*}[!t]\small
\vspace{-3mm}
  \caption{Anomaly detection performance for the texture categories of the MVTec AD dataset. For each cell in the ``$R_1$ / $R_2$'' columns, the ratio of correctly classified samples of normal $R_1$ and that of anomalous images $R_2$ are shown with ``$R_1$ / $R_2$'' notation.  The maximum averages $(R_1 + R_2) / 2$ are marked with boldface. The performance for the non-sparse-coding-based methods are cited from Table~2 of Bergmann et al.~\cite{mvtec_add}. The ``AUROC'' columns show only sparse coding and MLF-SC.}
  \label{tab:auc}
  \centering
  \setlength{\tabcolsep}{4pt}
  \begin{tabular}{ccccccccccc}
  \toprule
    & \multicolumn{7}{c}{$R_1$ / $R_2$} & & \multicolumn{2}{c}{AUROC} \\
    \cmidrule{2-8} \cmidrule{10-11}
     &  & & & CNN & Texture & Sparse & MLF-SC & & Sparse & MLF-SC\\
    Category & AE (SSIM) & AE (L2) & AnoGAN & Feature Dictionary & Inspection & Coding & (Proposed) & & Coding & (Proposed)\\ \midrule
    {\it Carpet} & 0.43 / 0.90 & 0.57 / 0.42 & 0.82 / 0.16 & 0.89 / 0.36 & 0.57 / 0.61 & 0.43 / 0.79 & {\bf 1.00} / {\bf 0.98} & & 0.58 & {\bf 0.99}\\
    {\it Grid} & 0.38 / 1.00 & 0.57 / 0.98 & 0.90 / 0.12 & 0.57 / 0.33 & 1.00 / 0.05 & 0.76 / 0.72 & {\bf 1.00} / {\bf 0.88} & & 0.89 & {\bf 0.97} \\
    {\it Leather} & 0.00 / 0.92 & 0.06 / 0.82 & 0.91 / 0.12 & 0.63 / 0.71 & 0.00 / 0.99 & 0.84 / 0.96 & {\bf 0.97} / {\bf 0.97} & & 0.95 & {\bf 0.99}\\
    {\it Tile} & 1.00 / 0.04 & 1.00 / 0.54 & 0.97 / 0.05 & 0.97 / 0.44 & 1.00 / 0.43 & 0.94 / 0.60 & {\bf 0.94} / {\bf 0.76} & & 0.86 & {\bf 0.92} \\
    {\it Wood} & 0.84 / 0.82 & 1.00 / 0.47 & 0.89 / 0.47 & 0.79 / 0.88 & 0.42 / 1.00 & 0.84 / 1.00 & {\bf 0.95} / {\bf 0.98} & & 0.97 & {\bf 0.99}\\
    \midrule
    Average & 0.53 / 0.74 & 0.64 / 0.65 & 0.90 / 0.18 & 0.77 / 0.54 & 0.60 / 0.62 & 0.76 / 0.81 & {\bf 0.97} / {\bf 0.91} & & 0.85 & {\bf 0.97}\\
    \bottomrule
  \end{tabular}
\end{table*}


We define the anomaly score of a given image to be the sum of the top-5 largest reconstruction errors from all the feature patches. In this way, we discard small error regions and focus only on (possibly) anomalous regions. This is because even an anomalous image contains many normal regions. If we sum the reconstruction errors of all the patches, an anomalous image with a small anomaly will be wrongly classified as normal. Therefore, by using only the top-5 errors, we can detect an anomaly even if its size is small.

\section{Experimental Results}
\label{sec:experimental_results}

We compare the performance of MLF-SC with traditional sparse coding and other state-of-the-art methods.

In our experiments, we use the five texture categories of the MVTec AD dataset~\cite{mvtec_add}, i.e.\ {\it Carpet}, {\it Grid}, {\it Leather}, {\it Tile}, and {\it Wood}. Training data and test data are separately contained in different folders.  All the training images do not contain anomalies. The test images are labeled normal or anomalous, and if an image is anomalous, the type of anomaly is also supplied (e.g., ``hole'' and ``cut'' for {\it Carpet} images). 

A dictionary for sparse coding was constructed for each category using the training data.
We used the following parameters for traditional sparse coding (on the original image): Patch size is 16${}\times{}$16, stride is 4, the number of bases is 200, and $\alpha$ in \eqref{eq:l1_sparse_coding} is 1. We used the following parameters for MLF-SC (on the multi-scale features): Patch size is 8${}\times{}$8, stride is 2, the number of bases is 5, and $\alpha$ in \eqref{eq:l1_sparse_coding} is 1.
Lasso-LARS~\cite{lars} is employed for optimization. The source code used for our experiments is available at \url{https://github.com/LeapMind/MLF-SC}.

The anomaly detection performance was measured with the test data. Evaluation metrics were the ratio of correctly classified normal images and the ratio of correctly classified anomalous images~\cite{mvtec_add} and the area under the receiver operation characteristics curve (AUROC). We compared the proposed method with the following six methods: autoencoder with SSIM loss, autoencoder  with L2 loss), AnoGan, CNN feature dictionary, texture inspection, and sparse coding.


Table~\ref{tab:auc} shows the performance of the seven methods for the texture datasets from MVTec AD, where MLF-SC always outperforms all the other methods. MLF-SC performs very good for all the five categories, suggesting the robustness of MLF-SC to many different types of textures, whereas other methods perform good only for some categories.


Fig.~\ref{fig:example_images_and_histograms_for_discussion} shows examples of normal and anomalous images from {\it Carpet} of MVTec AD (upper row) and the histograms of the reconstruction errors in the input layer, shallow layer, and deep layer (bottom row). The reconstruction errors of the four anomalous images in Fig.~\ref{fig:example_images_and_histograms_for_discussion}(a)--(d) are marked in the histograms. The leftmost histogram show that sparse coding at the input layer (original image) fails to detect anomalies because the reconstruction errors of the anomalous images are indistinguishable from those of normal images. The reconstruction errors of the anomalous images go out of those of the normal images as the layer goes deeper.


\section{Conclusion}
\label{sec:conclusion}

We have described how to incorporate multi-scale features to anomaly detection with sparse coding by using intermediate layers of a deep neural network, motivated by the experimental findings presented in Fig.~\ref{fig:when_sc_fails}.
It has been shown that the proposed method, MLF-SC, achieved the best anomaly detection performance among the state-of-the-art methods, thanks to its ability to cope of anomalies at different scales.

\bibliographystyle{IEEEtran}
\bibliography{refs}

\end{document}